\documentclass[lettersize,journal]{IEEEtran}
\usepackage{amsmath,amsfonts}
\usepackage{algorithmic}
\usepackage{algorithm}
\usepackage{array}
\usepackage[caption=false,font=normalsize,labelfont=sf,textfont=sf]{subfig}
\usepackage{textcomp}
\usepackage{stfloats}
\usepackage{url}
\usepackage{verbatim}
\usepackage{graphicx}
\usepackage{cite}
\usepackage{multirow}
\newcommand{\RNum}[1]{\uppercase\expandafter{\romannumeral #1\relax}}
\usepackage{array}
\newcommand{\mathbit}[1]{\mathbf{#1}}

\begin{document}

\title{Dynamic Topology Awareness: Breaking the\\ Granularity Rigidity in Vision-Language Navigation}

\author{Jiankun Peng, Jianyuan Guo, Ying Xu, Yue Liu, Jiashuang Yan, Xuanwei Ye, Houhua Li, Xiaoming Wang

    \thanks{Corresponding author: Ying Xu.}
    \thanks{Jiankun Peng, Yue Liu, Jiashuang Yan, Xuanwei Ye, Houhua Li is with the Aerospace Information Research Institute, Chinese Academy of Sciences, Beijing 100094, China; and is also with the School of Electronic, Electrical and Communication Engineering, University of Chinese Academy of Sciences, Beijing 100049, China (e-mail: pengjiankun24@mails.ucas.ac.cn; liuyue212@mails.ucas.ac.cn; yanjiashuang22@mails.ucas.ac.cn; yexuanwei24@mails.ucas.ac.cn; lihouhua24@mails.ucas.ac.cn).}
    \thanks{Jianyuan Guo is with the Department of Computer Science, City University of Hong Kong, Hong Kong, SAR, China (e-mail: jianyguo@cityu.edu.hk)}
    \thanks{Ying Xu and Xiaoming Wang are with the Aerospace Information Research Institute, Chinese Academy of Sciences, Beijing 100094, China (e-mail: xuying@aircas.ac.cn; wxm@aircas.ac.cn).}
 
}

\maketitle

\begin{abstract}
Vision-Language Navigation in Continuous Environments (VLN-CE) presents a core challenge: grounding high-level linguistic instructions into precise, safe, and long-horizon spatial actions. Explicit topological maps have proven to be a vital solution for providing robust spatial memory in such tasks. However, existing topological planning methods suffer from a “Granularity Rigidity” problem. Specifically, these methods typically rely on fixed geometric thresholds to sample nodes, which fails to adapt to varying environmental complexities. This rigidity leads to a critical mismatch: the model tends to over-sample in simple areas, causing computational redundancy, while under-sampling in high-uncertainty regions, increasing collision risks and compromising precision. To address this, we propose DGNav, a framework for Dynamic Topological Navigation, introducing a context-aware mechanism to modulate map density and connectivity on-the-fly. Our approach comprises two core innovations: (1) A Scene-Aware Adaptive Strategy that dynamically modulates graph construction thresholds based on the dispersion of predicted waypoints, enabling “densification on demand” in challenging environments; (2) A Dynamic Graph Transformer that reconstructs graph connectivity by fusing visual, linguistic, and geometric cues into dynamic edge weights, enabling the agent to filter out topological noise and enhancing instruction adherence. Extensive experiments on the R2R-CE and RxR-CE benchmarks demonstrate DGNav exhibits superior navigation performance and strong generalization capabilities. Furthermore, ablation studies confirm that our framework achieves an optimal trade-off between navigation efficiency and safe exploration. The code is available at https://github.com/shannanshouyin/DGNav.
\end{abstract}


\section{Introduction}
\IEEEPARstart{V}{ision-Language} Navigation (VLN) stands as a pivotal task in embodied AI, requiring agents to interpret natural language instructions and navigate through unseen environments to reach target locations \cite{anderson2018vision}. As the research focus shifts from discrete graphs to continuous environments (VLN-CE) \cite{krantz2020beyond}, agents face escalated challenges: they must not only perform high-level semantic reasoning but also execute low-level precise control \cite{xu2025joint}, \cite{liu2025joint}. In such complex tasks, endowing agents with explicit spatial memory and robust planning capabilities is imperative for success.

Current approaches typically fall into two paradigms. End-to-End learning methods directly map raw visual observations and instructions to actions \cite{zhang2024navid}, \cite{zhang2024uni}, \cite{wei2025streamvln}. While effective in short-horizon tasks, these “black-box” models often incur prohibitive computational costs and struggle to guarantee safety due to the lack of explicit geometric constraints \cite{krantz2020beyond}, \cite{raychaudhuri2021language}, \cite{krantz2021waypoint}. In contrast, Explicit Map-based approaches have emerged as a robust alternative \cite{hong2022bridging}, \cite{krantz2022sim}, \cite{an2024etpnav}, \cite{chen2022think}. By constructing a structured topological representation of the environment, these methods decouple high-level planning from low-level control. This modular hierarchy not only provides agents with interpretable spatial memory for long-horizon navigation but also imposes physical constraints necessary for precise obstacle avoidance.

Despite their effectiveness, existing topological planning methods are constrained by a fundamental bottleneck we term “Granularity Rigidity”. Most state-of-the-art methods rely on environment-agnostic thresholding rules to structure the agent's spatial memory and environmental perception. Specifically, these approaches predominantly utilize static geometric distances to determine the bias matrix of the Graph Transformer \cite{an2024etpnav}, \cite{chen2022think}, \cite{yue2024safe}, thereby dominating the edge connectivity weights regardless of the actual scene complexity. This reliance on fixed heuristics creates a structural decoupling between the map representation and the actual navigational complexity, manifesting in two critical limitations: First, the physical granularity is non-adaptive to environmental entropy. In structurally simple, low-uncertainty regions (e.g., straight corridors), fixed thresholds often result in node redundancy, introducing unnecessary computational noise. Conversely, in cluttered or open areas with high navigational uncertainty, the resulting graph becomes catastrophically sparse, failing to provide sufficient candidate waypoints for precise local maneuvering. Second, and more critically, the reliance on static geometry leads to “Navigational Myopia” \cite{chen2022think}, \cite{fang2019scene}, \cite{georgakis2022cross}, \cite{wen2025ovl}. Because the bias terms in conventional graph transformers encode only static geometric distance information, the model’s ability to attend to the semantic attributes of graph nodes is weakened, causing the planner to exhibit locally greedy behaviors. For instance, when commanded to “walk down the hall and take the second left”, the agent often assigns excessive attention to the first encountered door purely due to its physical proximity, resulting in premature turns and task failure. In essence, geometric proximity does not inherently imply navigational correctness. Yet, existing rigid topologies lack the “semantic elasticity” required to dynamically reconstruct graph connectivity in alignment with complex, multi-step linguistic instructions.

To break this rigidity, we propose DGNav (Dynamic Graph Navigation), a unified framework that introduces Dynamic Topology Awareness to enable flexible and context-sensitive navigation. DGNav innovates on two levels. At the physical structure level, we devise a Scene-Aware Adaptive Strategy. By analyzing the dispersion of predicted waypoints, this strategy dynamically adjusts the graph construction threshold ($\gamma$), achieving an adaptive granularity that “maintains sparsity in low-uncertainty simple regions to enhance efficiency, while dynamically densifying in high-uncertainty open or complex areas to ensure precise coverage”. At the semantic logic level, we propose a Dynamic Graph Transformer. Unlike traditional planners that rely on static edges, our module fuses visual similarity, instruction relevance, and geometric priors to dynamically generate context-aware edge weights. This mechanism allows the agent to functionally “disconnect” nearby geometric noise (e.g., the incorrect first turn) and establish “semantic shortcuts” to distant, task-relevant landmarks, effectively mitigating the myopia problem.

Unlike prior graph-based planners that operate on a fixed topology or static edge priors, DGNav explicitly adapts both graph granularity and connectivity online based on scene uncertainty and instruction semantics. The main contributions of this work are summarized as follows: (1) We identify the “Granularity Rigidity” problem in VLN-CE and propose the DGNav framework. Central to this framework, we introduce a Scene-Aware Adaptive Strategy, which, for the first time, dynamically modulates graph construction thresholds based on waypoint uncertainty, effectively resolving the conflict between node sparsity in complex regions and computational redundancy in open areas. (2) We propose a Multi-modal Dynamic Graph Transformer. Unlike traditional static geometric connectivity, this module reconstructs edge weights by fusing visual semantics, instruction relevance, and geometric constraints, thereby suppressing topological noise and significantly enhancing instruction fidelity in path planning. (3) We conduct extensive experiments on both R2R-CE and RxR-CE benchmarks. Results demonstrate that DGNav establishes a new performance benchmark, exhibiting significant superiority in terms of navigation accuracy and efficiency compared to both end-to-end learning baselines and other explicit map-based approaches. Furthermore, we investigate multiple advanced training strategies to analyze the impact of implicit geometric decoupling on performance upper bounds, providing empirical insights for enhancing navigation robustness in complex scenarios.

\section{RELATED WORK}
\subsection{VLN in Continuous Environments}
Early Vision-Language Navigation research primarily operated in discrete environments where agents navigate between pre-defined nodes \cite{anderson2018vision}. To bridge the gap to real-world robotics, Krantz et al. \cite{krantz2020beyond}, \cite{gu2022vision} introduced VLN in Continuous Environments (VLN-CE), requiring agents to execute low-level controls in continuous spaces. Existing approaches generally fall into two categories: end-to-end and modular methods. End-to-end approaches directly map pixel inputs to actions, relying on implicit neural representations for reasoning \cite{hong2022bridging}, \cite{chen2021history}—a paradigm recently bolstered by Vision-Language Models (VLMs) \cite{zhang2024navid}, \cite{zhang2024uni}, \cite{wei2025streamvln}. However, regardless of the backbone capacity, these methods inherently function as “black boxes” lacking explicit spatial structures. Consequently, they struggle to guarantee navigational safety and long-horizon consistency \cite{yan2025araim}, as the absence of geometric constraints often leads to unstable trajectories and frequent collisions. Conversely, modular approaches \cite{raychaudhuri2021language}, \cite{krantz2021waypoint}, \cite{an2024etpnav} decompose the task into waypoint prediction and local planning, offering better interpretability and stability. Our work follows the modular paradigm, aiming to enhance the robustness of the spatial planning module.

\subsection{Topological Mapping for Navigation}
Explicit spatial memory is pivotal for long-horizon navigation \cite{cadena2017past}. Topological maps, which represent environments as graphs of nodes and edges, have become a dominant solution due to their compactness compared to metric maps. Chen et al. \cite{chen2022think}, \cite{9577409} explored the application of topological graphs in VLN-CE based on offline pre-constructed prior maps. Building on these foundations, ETPNav \cite{an2024etpnav} advanced the field by proposing an “evolving” strategy that generates candidate nodes online and merges them incrementally. Despite these advancements, ETPNav and similar followers \cite{yue2024safe}, \cite{liu2024vision} inherit a critical limitation: granularity rigidity. First, regarding graph construction, they typically rely on fixed distance thresholds to merge nodes, ignoring the varying complexity of local scenes. Second, regarding edge connectivity, they predominately utilize static geometric distances (Euclidean distance) to determine edge weights. As argued in Section \RNum{1}, this static geometric prior fails to capture semantic connectivity, often leading the agent to prioritize geometrically close but semantically irrelevant nodes \cite{wen2025ovl}, \cite{chaplot2020neural}, \cite{kwon2021visual}.

\subsection{Dynamic Graph Learning in Navigation}
Graph Neural Networks (GNN) \cite{scarselli2008graph} and Transformers \cite{vaswani2017attention} have revolutionized the processing of non-Euclidean data. In the context of deep learning, the Transformer architecture can be viewed as a Graph Attention Network (GAT) \cite{velivckovic2017graph} operating on a fully connected graph, where attention weights represent dynamic edge strengths. While GNN have been applied to VLN to model environment topology \cite{an2024etpnav}, \cite{chen2022think}, \cite{yue2024safe}, \cite{9577409}, most existing methods treat the graph structure as static or purely geometric, only updating node features. In contrast, our DGNav extends the concept of dynamic graph learning to the topological structure itself. We introduce a mechanism to dynamically generate context-aware edge weights by fusing visual, linguistic, and geometric cues. By doing so, our Dynamic Graph Transformer functions as a learnable reasoning module that filters out topological noise and reconstructs the graph connectivity based on navigation context, rather than relying solely on physical proximity.

\begin{figure*}
    \centering
    \includegraphics[width=0.95\linewidth]{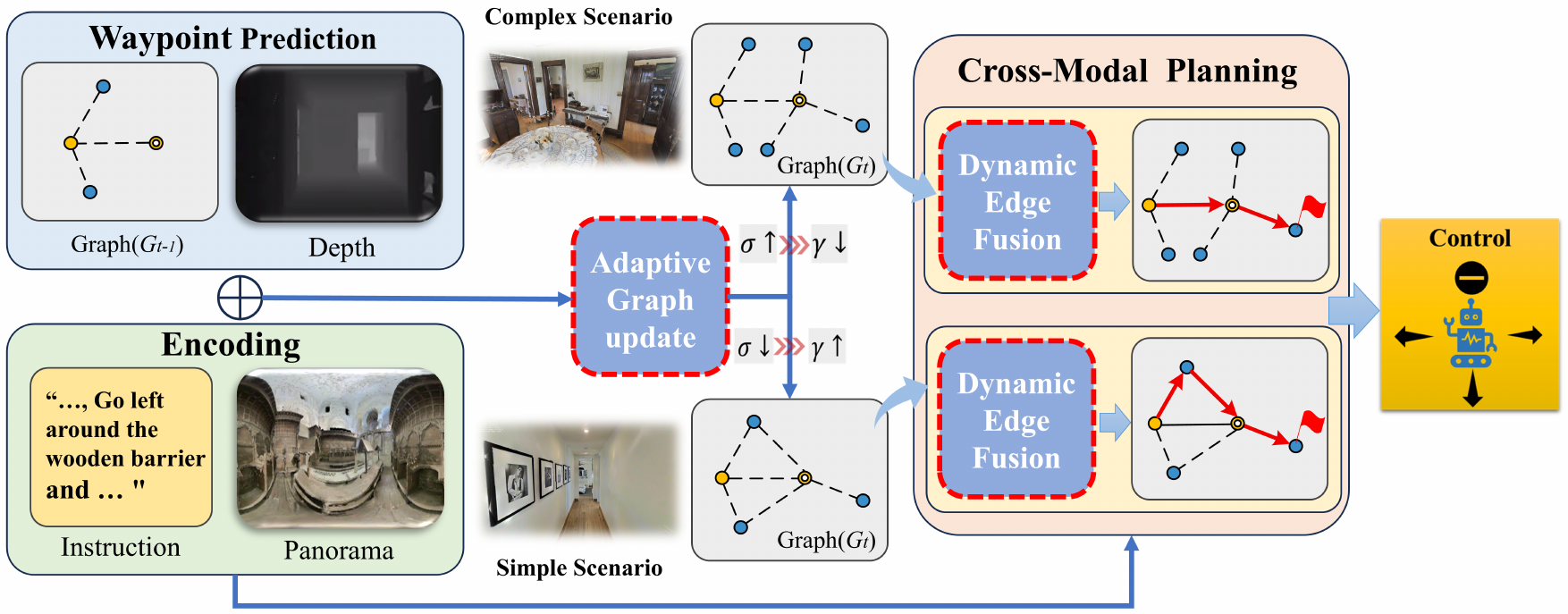}
    \caption{The overall framework of DGNav. DGNav dynamically adjusts its navigation strategy according to the estimated scene complexity $\sigma$. Specifically, higher scene complexity leads to denser topological graph construction, while simpler environments favor sparser representations. The graph merging threshold $\gamma$ controls the resulting graph granularity and is inversely correlated with $\sigma$, enabling adaptive trade-offs between navigation safety and efficiency.}
\end{figure*}

\section{METHOD}
\subsection{Problem Formulation and Overview}
The task of navigation in continuous environment is usually formulated as a sequential decision-making process. We address the VLN task in continuous environments. At step t, the agent receives a visual observation $O_t$ (RGB-D panorama) and a natural language instruction L. The goal is to predict a continuous action $a_t\in A$ (e.g., turning 15$^\circ$, moving forward 0.25 meters, and stopping) to reach the target. To handle long-horizon navigation, we maintain an online topological map $G_t = \langle v_t, \varepsilon_t \rangle$, where nodes $v_t$ represent navigable waypoints and edges $\varepsilon_t$ denote connectivity.

Following prior works in vision-and-language navigation \cite{krantz2021waypoint}, \cite{an2024etpnav}, \cite{chen2022think}, the agent incrementally constructs a topological graph to represent its explored environment. At each step, the agent leverages depth maps to infer navigable candidate locations, which are instantiated as graph nodes representing spatial positions, while panoramic RGB-D observations are used to extract semantic features associated with these nodes. Edges are then established between nodes based on geometric reachability, encoding feasible transitions for continuous navigation. As the agent moves, newly observed locations are added to the graph, while previously visited nodes are retained to provide a persistent spatial memory. This online topological representation enables the planner to reason over both local observations and accumulated global structure during long-horizon navigation.

Building upon this formulation, as illustrated in Fig.1, the proposed DGNav framework operates in a modular pipeline. First, the Waypoint Prediction module takes the depth map at step t and the topological graph from the previous step $G_{t-1}$ as inputs to predict candidate ghost nodes, which serve as potential navigable extensions of the current graph. These candidates are then processed by the Adaptive Graph Update module (our first innovation, detailed in Section \RNum{3}-B). This module analyzes the local scene complexity and adaptively regulates the graph granularity—generating a dense topology for complex scenarios to ensure safety, while maintaining a sparse topology for simple scenarios to enhance efficiency. Simultaneously, an Encoding module extracts feature representations from the RGB-D panorama and the natural language instruction. These encoded features, along with the updated graph $G_t$, are injected into the Cross-Modal Planning module. Inside this planner, the Dynamic Edge Fusion module (our second innovation, detailed in Section \RNum{3}-C) fuses multimodal cues—spanning static geometry, visual semantics, and natural language instructions—to construct a dynamic adjacency matrix $\mathbit{E}$, which is then consumed by the Cross-Modal Graph Transformer to reason about the optimal path. Finally, the high-level plan is executed by the Control module to output low-level actions.

\subsection{Scene-Aware Adaptive Topology}
\emph{\bfseries Waypoint Prediction and Granularity Definition}: As illustrated in Fig.2, our pipeline initiates with a Waypoint Prediction module, which perceives the local geometry by processing the depth map at step t and the topological graph from the previous step $G_{t-1}$. This module generates a set of candidates ghost nodes, denoted as $C_t=\ \left\{c_{t,i}\right\}_{i=1}^{N_c}$. Subsequently, the Adaptive Graph Update module determines the status of these candidates: whether to instantiate them as new landmarks or merge them into the existing graph. This decision is strictly governed by a distance threshold $\gamma$. Specifically, a candidate is merged if its Euclidean distance to the nearest neighbor is less than $\gamma$; otherwise, the graph is expanded. Thus, $\gamma$ serves as a critical controller for the granularity of environmental abstraction. A small $\gamma$ yields a dense topology (high-fidelity but computationally redundant), while a large $\gamma$ results in a sparse topology (efficient but potentially losing detail). Existing rigid methods typically fix $\gamma$, failing to accommodate varying scene requirements.

\emph{\bfseries Metric of Scene Complexity}: To break this rigidity, we leverage the spatial distribution of candidates $C_t$ as a proxy for local scene complexity. Specifically, given $N_c$ valid candidate nodes predicted at the current step, let $\theta_i$ denote the relative angle of the $i$-th candidate with respect to the agent's heading. We quantify the spatial dispersion $\sigma_t$ by calculating the standard deviation of these angles:

\begin{equation}
\label{deqn_ex1a}
\sigma_t=\sqrt{\frac{1}{N_c}\bullet\sum_{i=1}^{N_c}\left(\theta_i-\bar{\theta}\right)^\mathbf{2}}.
\end{equation}

where $\bar{\theta}$ represents the mean angle of all candidates. As depicted in the three branches of Fig. 2, a high $\sigma_t$ (e.g., at an intersection where candidates diverge) implies a complex decision boundary, while a low $\sigma_t$ (e.g., in a corridor where candidates cluster forward) suggests a simple geometric structure. Accordingly, we establish a linear inverse control law to dynamically modulate $\gamma_t$:

\begin{equation}
\label{deqn_ex1a}
\gamma_t=CLIP\left(\alpha-\beta\bullet\sigma_t,\ \gamma_{min},\gamma_{max}\right).
\end{equation}

This formulation ensures that in complex regions (high $\sigma_t$), $\gamma_t$ is decreased to construct a dense graph for safety, whereas in open areas (low $\sigma_t$), $\gamma_t$ is increased to form a sparse graph for efficiency.

\begin{figure}[!t]
\centering
\includegraphics[width=3.5in]{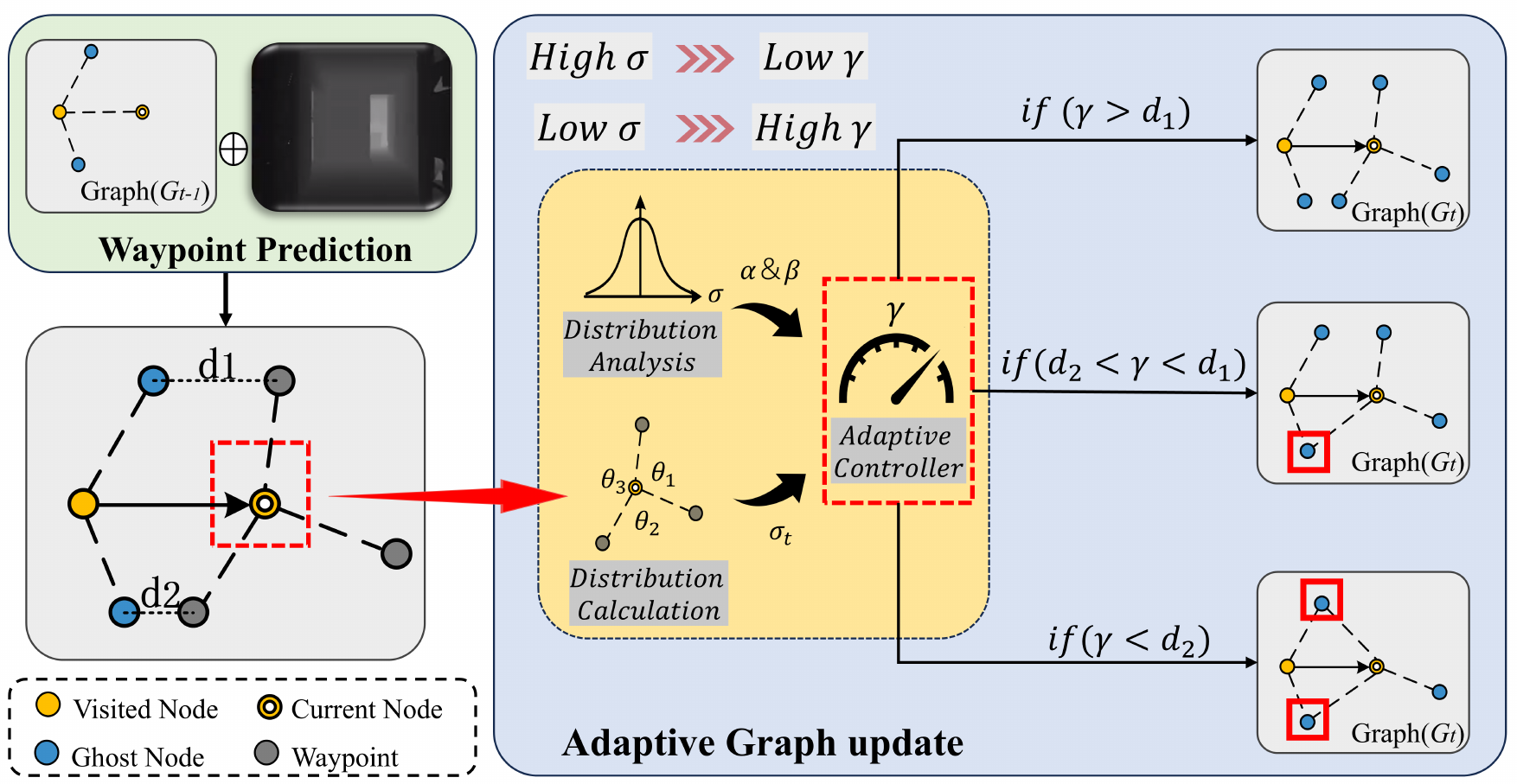}
\caption{Illustration of the Scene-Aware Adaptive Strategy. The process begins with Waypoint Prediction, generating raw ghost nodes from the depth map. These nodes are then passed to the Adaptive Graph Update module. Based on the angular dispersion ($\sigma$) of the candidates, the controller dynamically adjusts the merging threshold $\gamma$. The figure demonstrates three scenarios: in simple environments (low $\sigma$), a larger $\gamma$ yields a sparse graph for efficiency; in complex environments (high $\sigma$), a smaller $\gamma$ results in a dense graph for safety.}
\label{fig_2}
\end{figure}

\emph{\bfseries Statistical Calibration of Parameters}: The coefficients $\alpha$ and $\beta$ in our control law are not heuristically tuned hyperparameters but are derived through rigorous statistical calibration. Specifically, we utilized the fully converged ETPNav baseline model \cite{an2024etpnav} to perform inference across the Val-Seen split (i.e., a subset of instructions from the R2R-CE training set, used to strictly avoid data leakage from unseen environments), collecting the angular dispersion $\sigma_t$ of candidate ghost nodes at every navigation step. As illustrated in Fig. 3, we plot the statistical distribution histogram of $\sigma_t$. It can be observed that the distribution exhibits a distinct Normal Distribution (Gaussian-like) profile. This indicates that the complexity of most navigation scenarios clusters around the central mean, while extremely open or cluttered scenarios reside in the tails.

\emph{\bfseries Theoretical Justification for Linear Mapping}: Based on this Gaussian observation, we theoretically justify the choice of a linear control law over non-linear alternatives (e.g., Sigmoid or Exponential). From an information-theoretic perspective, a linear transformation f(x)=$\alpha-\beta x$ preserves the maximum entropy property of the Gaussian source distribution P($\sigma_t$) \cite{cover1999elements}, \cite{bishop2006pattern}. In contrast, non-linear mappings like Sigmoid introduce saturation regions at the distribution tails where gradients vanish ($\nabla f\rightarrow0$). This would compress distinct high-uncertainty states (the “cluttered” tail) into indistinguishable threshold values, causing information loss in the most critical scenarios. Therefore, the linear mapping acts as an optimal first-order approximation, maintaining constant sensitivity across the full dynamic range of scene complexity. To empirically validate this analysis, we provide a comparative study of different mapping functions in Section \RNum{4}-C, demonstrating the superior stability of our linear approach compared to non-linear variants.

\begin{figure}[!t]
\centering
\includegraphics[width=2.5in]{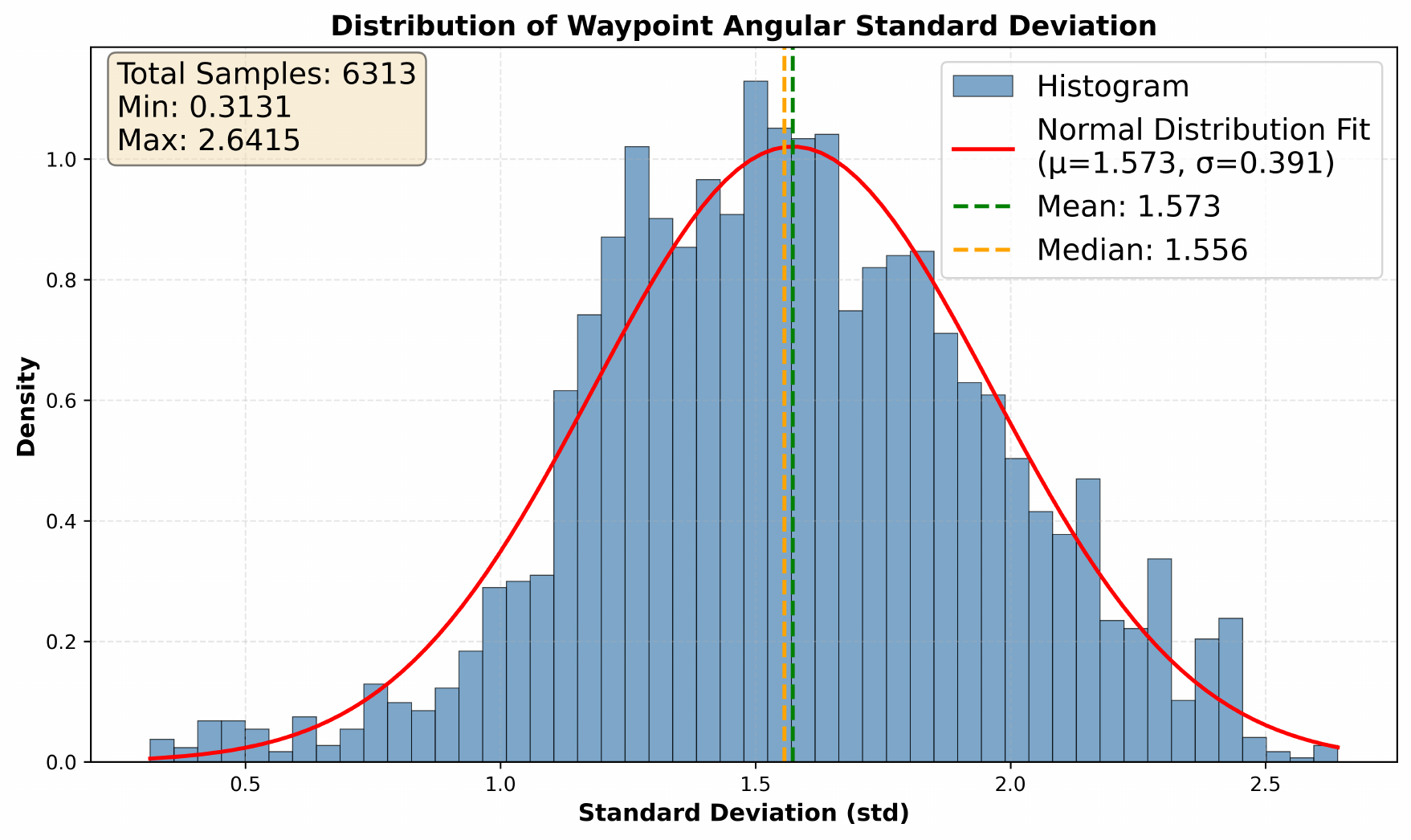}
\caption{Statistical distribution of angular dispersion ($\sigma_t$) for parameter calibration. Data is collected from the ETPNav baseline on the R2R-CE Val-Seen split. The distribution exhibits a Gaussian-like profile, which serves as the empirical basis for calibrating $\alpha$ and $\beta$.}
\label{fig_3}
\end{figure}

\emph{\bfseries Conditional Linear Mapping Strategy}: Applying dynamic adjustment across the entire uncertainty range may introduce unnecessary topological fluctuations in stable environments. To address this issue, we adopt a conditional control strategy that activates adaptive granularity only when scene complexity exceeds a critical level. Specifically, the median dispersion $\sigma_{med}$ is used to delineate a stable regime, within which the merging threshold is fixed at a conservative baseline $\gamma_{fix}$. When $\sigma_t>\sigma_{med}$, the threshold is progressively relaxed to allow finer topological refinement, with the adjustment bounded by the maximum dispersion $\sigma_{max}$. This design ensures that structural adaptation is triggered only in decision-critical regions while preserving stability in common cases. The specific mapping relationship is as follows:
\begin{equation}
\label{deqn_ex1a}
\gamma_t=\ \left\{\begin{matrix}\gamma_{fix}&if\ \sigma_t\ \le\ \sigma_{med}\\\gamma_{fix}-\left(\frac{\sigma_t-\sigma_{med}}{\sigma_{max}-\sigma_{med}}\right)\left(\gamma_{fix}-\gamma_{min}\right)&if\ \sigma_t\ >\ \sigma_{med}\\\end{matrix}\right.
\end{equation}

This formulation guarantees continuity at $\sigma_{med}$ while activating adaptive thresholding only in high-uncertainty regimes. Empirical validation in Section \RNum{4}-C, comparing global versus conditional linear mappings, confirms that the conditional approach more effectively captures scene complexity to attain higher navigation success rates.

\begin{figure}[!t]
\centering
\includegraphics[width=3.5in]{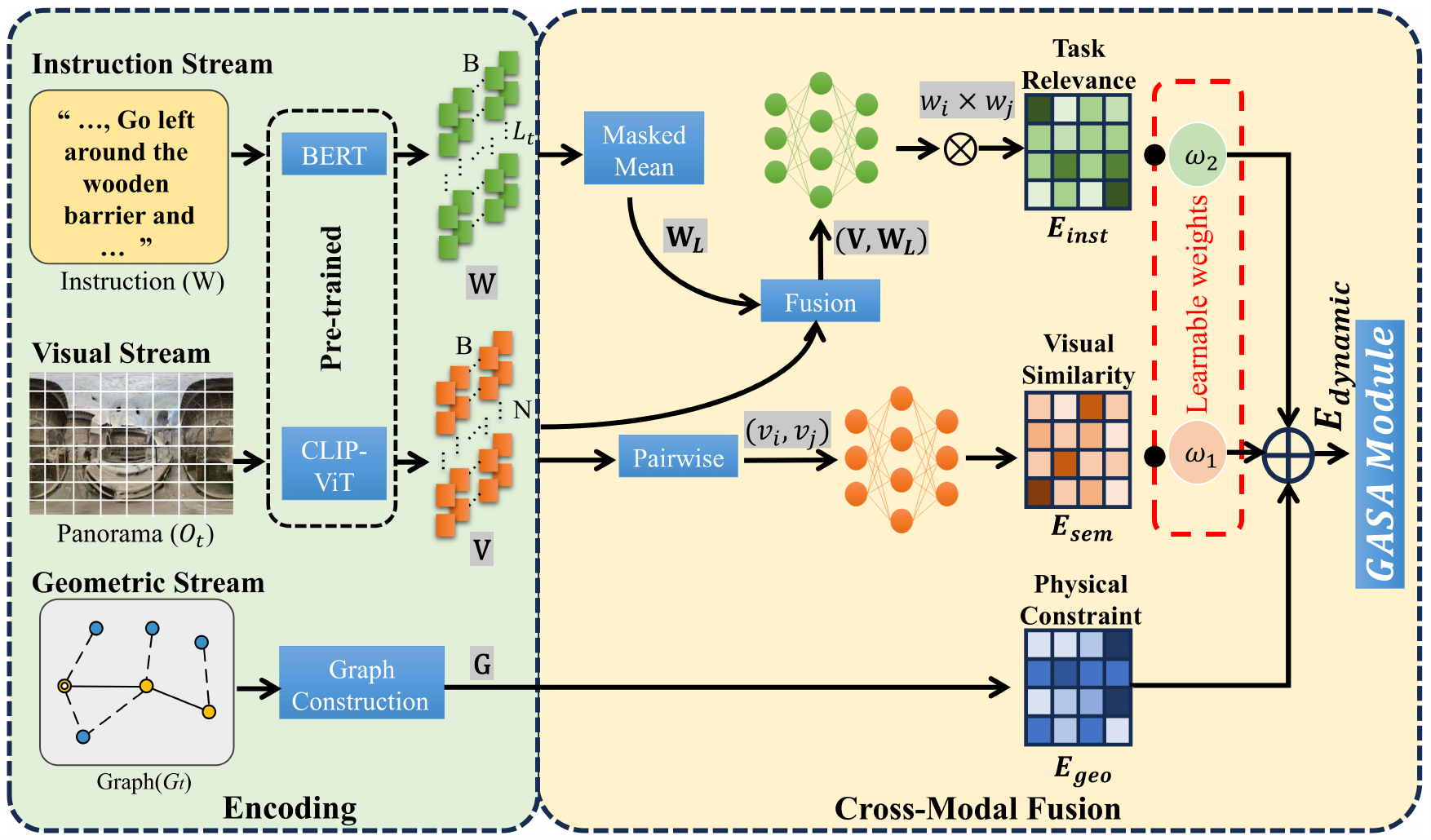}
\caption{The architecture of Multimodal Encoding and Dynamic Edge Fusion. The visual encoder and instruction encoder extract node features ($\mathbf{V}$) and word features ($\mathbf{W}$), respectively. The Dynamic Edge Fusion module constructs the graph connectivity by fusing the geometric map ($\mathbf{E}_{\mathbit{geo}}$), pairwise visual similarity ($\mathbf{E}_{\mathbit{sem}}$), and instruction relevance ($\mathbf{E}_{\mathbit{inst}}$ derived from $\mathbf{W}_\mathbit{L}$). The resulting matrix $\mathbf{E}_{\mathbit{dynamic}}$ guides the Graph Transformer to perform context-aware planning}
\label{fig_4}
\end{figure}

\subsection{Dynamic Graph Transformer}
\emph{\bfseries Feature Encoding}: As illustrated in Fig. 4, the system processes multimodal inputs into a unified feature space. For the linguistic stream, we employ a pre-trained BERT \cite{devlin2019bert} to encode instructions on R2R-CE, and RoBERTa \cite{liu2019roberta} for multilingual instruction encoding on RxR-CE, yielding the word feature sequence $\mathbf{W}\in\mathbb{R}^{L\times D}$, where $\ L$ is the instruction length. We further aggregate $\mathbf{W}$ to obtain the global instruction token $\mathbf{W}_\mathbit{L}\in\mathbb{R}^D$ via pooling. For the visual stream, the topological map consists of N navigable nodes. We extract the panoramic features of each node using a pre-trained CLIP-ViT \cite{radford2021learning}, \cite{dosovitskiy2020image}, fused with orientation embeddings, to generate the visual node features $\mathbf{V}=\left\{v_1,...,v_N\right\}\in\mathbb{R}^{N\times D}$.

\emph{\bfseries Dynamic Edge Fusion}: To enable semantic reasoning, we reconstruct the graph connectivity by fusing three distinct streams into a dynamic adjacency matrix $\mathbit{E}_{\mathbit{dynamic}}$.
\begin{itemize}
  \item \emph{Geometric Stream}: Provides the hard constraint of the physical world.
  We calculate the normalized Euclidean distance $\mathbit{E}_{\mathbit{geo}}^{\left(\mathbit{i},\mathbit{j}\right)}$, representing basic spatial reachability.
   \item \emph{Visual Stream}: Introduces semantic consistency to soften physical constraints. To capture semantic continuity, we compute the pairwise similarity between visual nodes $v_i$ and $v_j$. By passing them through a non-linear mapping, we derive the semantic adjacency matrix:
   \begin{equation}
    \label{deqn_ex1a}
    \mathbf{E}_{sem}^{\left(i,j\right)}=MLP\left(\left[v_i;v_j\right]\right).
    \end{equation}
    This term enables the model to connect regions that are visually coherent even if they are not immediate physical neighbors.
    \item \emph{Instruction Stream}: To align the topology with the task, we compute the relevance between each node $v_i$ and the global instruction $\mathbit{W}_\mathbit{L}$. The instruction-guided edge weight is derived via the outer product of relevance scores:
    \begin{equation}
    \label{deqn_ex1a}
    w_i=MLP\left(\left[v_i;\mathbf{W}_\mathbit{L}\right]\right).
    \end{equation}
    \begin{equation}
    \label{deqn_ex1a}
    \mathbf{E}_{inst}^{\left(i,j\right)}=w_i\bullet w_j.
    \end{equation}
    This term effectively filters task-irrelevant topological noise by suppressing paths that do not align with the instruction description.
\end{itemize}

The final dynamic adjacency matrix $\ \mathbit{E}_{\mathbit{dynamic}}$ is formulated by superimposing learnable semantic residuals onto the fixed physical baseline:
\begin{equation}
\label{deqn_ex1a}
\mathbf{E}_{\mathbit{dynamic}}=\mathbf{E}_{\mathbit{geo}}+\omega_1\bullet\mathbf{E}_{\mathbit{sem}}+\omega_2\bullet\mathbf{E}_{\mathbit{inst}}.
\end{equation}

Here, $\omega_1$ and $\omega_2$ are learnable scalar coefficients. This architecture allows the model to smoothly transition from pure geometric navigation to semantic-aware planning, preventing instability during the early stages of training.

\emph{\bfseries Graph Transformer with Dynamic Bias}: The constructed dynamic adjacency matrix $\mathbf{E}_{\mathbit{dynamic}}$ serves as a learnable structural bias for the graph reasoning module. We employ a multi-layer Graph Transformer to update the node features. Let ${\hat{\mathbf{H}}}^\mathbit{l}=\left\{h_1^l,...,h_N^l\right\}$ denote the node features at layer$\ l$. The update process for the $(l+1)$-th layer is formalized as:
\begin{equation}
\label{deqn_ex1a}
{\hat{\mathbf{H}}}^{\mathbit{l}+\mathbf{1}}=GASA\left(\mathbf{H}^\mathbit{l},\mathbf{E}_{\mathbit{dynamic}}\right)+\mathbf{H}^\mathbit{l}.
\end{equation}
\begin{equation}
\label{deqn_ex1a}
\mathbf{H}^{\mathbit{l}+\mathbf{1}}=FFN\left(LN\left({\hat{\mathbf{H}}}^{\mathbit{l}+\mathbf{1}}\right)\right)+{\hat{\mathbit{H}}}^{\mathbit{l}+\mathbf{1}}.
\end{equation}

where FFN denotes the Feed-Forward Network and LN denotes Layer Normalization. Specifically, the Graph-Aware Self-Attention (GASA) mechanism \cite{an2024etpnav}, \cite{chen2022think}, \cite{scarselli2008graph}, \cite{velivckovic2017graph} incorporates the dynamic topology directly into the attention calculation. For a single attention head, the output is computed as:

\begin{equation}
\label{deqn_ex10} 
\begin{split}
\mathrm{GASA}\!\left(\mathbf{H}^l, \mathbf{E}_{\mathrm{dynamic}}\right)
&= \mathrm{Softmax}\!\left(
\frac{
\mathbf{H}^l \mathbf{W}_Q
\left(\mathbf{H}^l \mathbf{W}_K\right)^{\mathsf{T}}
}{
\sqrt{d_k}
} \right. \\ 
&\quad \left. + \mathbf{E}_{\mathrm{dynamic}}
\right) 
\left(\mathbf{H}^l \mathbf{W}_V\right).
\end{split}
\end{equation}

Here, $\mathbit{W}_\mathbit{Q}, \mathbit{W}_\mathbit{K}, \mathbit{W}_\mathbit{v}$ are learnable projection matrices. By replacing the static geometric mask with our $\mathbit{E}_{\mathbit{dynamic}}$, the attention mechanism is forced to attend to nodes that are semantically relevant ($\omega_1\bullet\mathbf{E}_{\mathbit{sem}}$) and task-aligned ($\omega_2\bullet\mathbf{E}_{\mathbit{inst}}$), effectively creating "soft" semantic edges that bypass the rigid constraints of the physical graph

\begin{table*}[t]
\centering
\caption{END-TO-END VS. EXPLICIT MAP-BASED APPROACHES}
\label{tab:comparison}
\setlength{\extrarowheight}{2pt}
\resizebox{\textwidth}{!}{%
\begin{tabular}{c|c|ccccc|ccccc|ccccc}
\hline

\noalign{\hrule height 0.1pt}
\multicolumn{2}{c}{\multirow{2}{*}{Methods}} & \multicolumn{5}{c}{Val-Seen} & \multicolumn{5}{c}{Val-Unseen} & \multicolumn{5}{c}{Test-Unseen} \\ \cline{3-17} 
\multicolumn{2}{c}{} & TL$\downarrow$ & NE$\downarrow$ & OSR$\uparrow$ & SR$\uparrow$ & SPL$\uparrow$ & TL$\downarrow$ & NE$\downarrow$ & OSR$\uparrow$ & SR$\uparrow$ & SPL$\uparrow$ & TL$\downarrow$ & NE$\downarrow$ & OSR$\uparrow$ & SR$\uparrow$ & SPL$\uparrow$ \\ \hline

\multirow{7}{*}{\shortstack{End-to-\\End}} 
 & Seq2Seq\cite{chang2017matterport3d} & 9.26 & 7.12 & 46 & 37 & 35 & 8.64 & 7.37 & 40 & 32 & 30 & 8.85 & 7.91 & 36 & 28 & 25 \\
 & Sim2Sim\cite{krantz2022sim} & 11.18 & 4.67 & 61 & 52 & 44 & 10.69 & 6.07 & 52 & 43 & 36 & 11.43 & 6.17 & 52 & 44 & 37 \\
 & CWP-CMA\cite{hong2022bridging} & 11.47 & 5.20 & 61 & 51 & 45 & 10.90 & 6.20 & 52 & 41 & 36 & 11.85 & 6.30 & 49 & 38 & 33 \\
 & CWP-RecBert\cite{hong2022bridging} & 12.50 & 5.02 & 59 & 50 & 44 & 12.23 & 5.74 & 53 & 44 & 39 & 13.51 & 5.89 & 51 & 42 & 36 \\
 & Navid\cite{zhang2024navid} & - & - & - & - & - & 7.63 & 5.47 & 49 & 37 & 36 & - & - & - & - & - \\
 & Uni-Navid\cite{zhang2024uni} & - & - & - & - & - & 9.71 & 5.58 & 53 & 47 & 43 & - & - & - & - & - \\
 & StreamVLN\cite{wei2025streamvln} & - & - & - & - & - & - & 5.43 & 63 & 53 & 47 & - & - & - & - & - \\ \hline

\multirow{8}{*}{\shortstack{Explicit\\Map-\\based}} 
 & CM2\cite{georgakis2022cross} & 12.05 & 6.10 & 50.7 & 42.9 & 34.8 & 11.54 & 7.02 & 41.5 & 34.3 & 27.6 & 13.9 & 7.7 & 39 & 31 & 24 \\
 & WG-MGMap\cite{chen2022weakly} & 10.12 & 5.65 & 52 & 47 & 43 & 10.00 & 6.28 & 48 & 39 & 34 & 12.30 & 7.11 & 45 & 35 & 28 \\
 & DUET-CE\cite{chen2022think} & - & - & - & - & - & 13.08 & 5.16 & 62 & 54 & 46 & - & - & - & - & - \\
 & GridMM\cite{wang2023gridmm} & 12.69 & 4.21 & 69 & 59 & 51 & 13.36 & 5.11 & 61 & 49 & 41 & 13.31 & 5.54 & 56 & 46 & 39 \\
 & Safe-VLN\cite{yue2024safe} & 13.71 & 3.35 & 79 & 71 & 60 & 15.00 & 4.48 & 68 & 60 & 47 & 15.44 & 5.01 & 54 & 56 & 45 \\
 & OVL-MAP\cite{wen2025ovl} & 11.82 & 3.90 & 73 & 68 & 59 & 11.45 & 4.69 & 65 & 58 & 50 & 11.64 & 4.98 & 62 & 57 & 48 \\
 & ETPNav\cite{an2024etpnav} & 11.78 & 3.95 & 72 & 66 & 59 & 11.99 & 4.71 & 65 & 57 & 49 & 12.87 & 5.12 & 63 & 55 & 48 \\ \cline{2-17} 
 & \textbf{Ours (DGNav)} & 10.79 & 3.78 & 72 & 66 & \textbf{60} & 11.64 & 4.66 & 65 & 59 & \textbf{50} & 14.20 & \textbf{4.92} & \textbf{64} & 56 & 47 \\ \hline

\noalign{\hrule height 0.1pt}
\end{tabular}%
}
\end{table*}

\section{EXPERIMENTS}
\subsection{Experimental Setup}
\emph{\bfseries Datasets}: We evaluate our method on two standard Vision-Language Navigation datasets in continuous environments (VLN-CE), i.e., R2R-CE and RxR-CE,, both built upon the Matterport3D (MP3D) \cite{anderson2018vision} simulation platform. MP3D serves as the visual backbone, providing 90 diverse indoor building environments with high-fidelity RGB-D scans. Unlike discrete navigation settings, the VLN-CE \cite{krantz2020beyond} benchmark allows agents to move freely in 3D space without being constrained to pre-defined graph nodes, imposing stricter requirements on spatial awareness and obstacle avoidance. (1) R2R-CE: This dataset is a continuous reconstruction of the Room-to-Room (R2R) benchmark. It contains 5611 trajectories across 90 scenes, paired with human-annotated English instructions. The dataset is split into Training, Val-Seen, Val-Unseen, and Test sets. The Val-Unseen and Test splits, which involve environments not encountered during training, serve as the primary benchmarks for evaluating the generalization capability of our method. (2) RxR-CE: To further verify the robustness of DGNav in complex, long-horizon tasks, we also conduct experiments on the Room-across-Room (RxR) continuous benchmark \cite{ku2020room}. Compared to R2R-CE, RxR-CE features significantly longer path trajectories, more fine-grained instruction descriptions, and a larger scale of instructions (over 10$\times$ that of R2R), covering multiple languages (English, Hindi, and Telugu). This dataset presents a more challenging testbed, demanding superior topological memory capabilities over extended time horizons.

\emph{\bfseries Evaluation Metrics}: To comprehensively assess navigation performance, we adopt the standard evaluation protocols for VLN-CE \cite{zhu2021deep}, \cite{xiong2025sensing}. We report the following metrics: (1) Basic Navigation Metrics: Trajectory Length (TL), Navigation Error (NE), Success Rate (SR), and Oracle Success Rate (OSR).  (2) Efficiency and Fidelity Metrics: Success weighted by Path Length (SPL), which balances success with path efficiency, along with Normalized Dynamic Time Warping (nDTW) and Spatial Dynamic Time Warping (SDTW), which evaluate the spatiotemporal alignment between the predicted and ground-truth trajectories. 

Following community conventions \cite{anderson2018evaluation}, \cite{ilharco2019general}, for the R2R-CE benchmark, we prioritize SR and SPL as the primary indicators of goal-reaching capability. For the RxR-CE benchmark, given its strict instruction-following requirements, nDTW and SDTW are considered the most critical metrics for measuring the agent's fidelity to the described path.

\emph{\bfseries Implementation Details}:

1)\emph{Training and Optimization}: We implemented our model using PyTorch and conducted all experiments in the Habitat simulator \cite{9010745} on two NVIDIA RTX 4090 GPUs. Following the established training protocol of ETPNav \cite{an2024etpnav}, we adopt a two-stage training strategy. First, we initialize the ViT \cite{dosovitskiy2020image} and BERT \cite{devlin2019bert}, \cite{liu2019roberta} using weights pre-trained on large-scale datasets. Second, we fine-tune the entire model on the R2R-CE and RxR-CE datasets. For R2R-CE, we employ the AdamW optimizer with a learning rate of 1e-5, a batch size of 8 (4 per GPU), and train for 20,000 iterations. For RxR-CE, we train for 25,000 iterations with the same learning rate of 1e-5.

2)\emph{Dynamic Graph Parameters}: For the dynamic graph module, the weight of the geometric stream is fixed at $\omega_1=1.0$. The learnable semantic weights, $\omega_2$ and $\omega_3$, are initialized to 0.1, and their specific learning rate is set to 1e-4 to facilitate rapid convergence from scratch.

3)\emph{Adaptive Strategy}: To ensure the stability of topological learning, we keep $\gamma$ fixed at the baseline value (0.5m) \cite{an2024etpnav} during the training phase. The Scene-Aware Adaptive Strategy is activated exclusively during the inference phase. At this stage, the merging threshold $\gamma$ is dynamically modulated between 0.25m and 0.5m based on the real-time $\sigma_t$. This decoupled strategy ensures stable feature representation learning while endowing the agent with the flexibility to handle extreme scenarios in unseen test environments. This adaptive mechanism does not rely on any additional supervision or privileged information during inference.

\subsection{Comparison with Other Methods}

Table I presents the performance comparison between DGNav and a series of classic navigation baselines. To provide a clear contextual analysis, we categorize baselines into two distinct paradigms: End-to-End Learning Methods (ranging from classic sequential models to recent VLM-based agents) and Explicit Map-based Methods (including topological, grid-based, Semantic planners).

\emph{\bfseries Comparison with End-to-End Learning Methods}: We first compare DGNav with some classic end-to-end baselines. As shown in Table I, these methods, which rely on implicit memory states, generally suffer from poor long-horizon navigability. DGNav achieves substantial superiority over these baselines, demonstrating the decisive advantage of our hierarchical planning framework that decouples perception from control. Furthermore, compared to recent Large Vision-Language Models (VLMs) based approaches such as NaVid \cite{zhang2024navid} and Uni-NaVid \cite{zhang2024uni}, DGNav maintains a strong competitive edge.

While these VLM-based agents achieve breakthroughs in zero-shot reasoning, their lack of explicit geometric constraints often leads to navigational instability in continuous environments. DGNav significantly outperforms Uni-NaVid by a large margin, confirming that despite the rise of large models, explicit spatial memory remains a crucial component for ensuring precise and robust navigation.

\emph{\bfseries Comparison with Explicit Map-based Methods}: Within the map-based category, DGNav establishes a new benchmark for topological planning. Versus Classic Map Baselines: Compared to early metric map approaches, DGNav exhibits substantial performance gains across all metrics, validating the efficacy of our dynamic graph construction over static mapping strategies.

When benchmarked against recent advanced baselines, DGNav demonstrates distinct advantages arising from its adaptive mechanism. Specifically, in comparison with GridMM \cite{wang2023gridmm}, which employs fine-grained grid maps, DGNav achieves significantly lower NE, suggesting that our adaptive topology offers greater flexibility for precise localization than rigid grid representations. Furthermore, while DGNav trails marginally behind Safe-VLN \cite{yue2024safe} in terms of SR and NE, it achieves a significantly shorter TL and substantially outperforms Safe-VLN in SPL on the Val-unseen split. This trade-off indicates that DGNav exhibits more intelligent behavior: rather than adopting overly conservative strategies that prolong paths, our method ensures high success rates while maintaining superior efficiency, reflecting a deeper understanding of both environmental scenes and linguistic instructions. Finally, even against OVL-Map \cite{wen2025ovl}, a formidable competitor incorporating object-level semantic mapping, DGNav maintains a crucial edge in SR (+1\%) and NE (-0.03m). This finding highlights that static accumulation of semantic labels is insufficient; instead, DGNav’s “Dynamic Edge Fusion” mechanism intelligently filters and leverages semantic cues based on instruction requirements, thereby enabling optimal navigation decisions in complex, unseen environments.

\begin{table*}[t]
\centering
\caption{COMPARISON WITH CLASSIC BASELINES ON RXR-CE}
\label{tab:table2}
\setlength{\tabcolsep}{10pt} 
\renewcommand{\arraystretch}{1.2} 
\begin{tabular}{c|ccccc|ccccc}
\noalign{\hrule height 0.8pt} 

\multicolumn{1}{c}{\multirow{2}{*}{Methods}} & \multicolumn{5}{c}{Val-Seen} & \multicolumn{5}{c}{Val-Unseen} \\ \cline{2-11} 

\multicolumn{1}{c}{} & NE$\downarrow$ & SR$\uparrow$ & SPL$\uparrow$ & nDTW$\uparrow$ & \multicolumn{1}{c|}{SDTW$\uparrow$} & NE$\downarrow$ & SR$\uparrow$ & SPL$\uparrow$ & nDTW$\uparrow$ & SDTW$\uparrow$ \\ \hline

Seq2Seq\cite{chang2017matterport3d} & - & - & - & - & - & 12.10 & 13.9 & 11.9 & 30.8 & - \\
CWP-CMA\cite{hong2022bridging} & - & - & - & - & - & 8.76 & 26.59 & 22.16 & 47.05 & - \\
CWP-RecBert\cite{hong2022bridging} & - & - & - & - & - & 8.98 & 27.08 & 22.65 & 46.71 & - \\ \hline
StreamVLN\cite{wei2025streamvln} & - & - & - & - & - & 6.72 & 48.6 & 42.5 & 60.2 & - \\
ETPNav\cite{an2024etpnav} & 5.55 & 57.26 & 47.67 & 64.15 & 47.57 & 5.80 & 53.07 & 44.16 & 61.49 & 43.92 \\
\textbf{Ours (DGNav)} & \textbf{5.43} & \textbf{58.48} & \textbf{48.54} & \textbf{65.49} & \textbf{48.60} & 6.00 & \textbf{53.78} & \textbf{44.37} & \textbf{62.04} & \textbf{44.49} \\ 

\noalign{\hrule height 0.8pt} 
\end{tabular}
\end{table*}

Finally, compared to our direct baseline ETPNav, DGNav achieves comprehensive improvements across NE, OSR, SR, and SPL on the R2R-CE Val-unseen split. On the more challenging Test-unseen split, while SPL shows a slight decrease (-1\%), DGNav demonstrates enhanced robustness with improvements in NE (-0.2m), OSR (+1\%), and SR (+1\%). This specific trade-off reflects DGNav's superior robustness: when facing highly unfamiliar and complex test environments, the adaptive graph mechanism encourages the agent to adopt a more cautious exploration strategy (e.g., moving slightly more to verify landmarks in ambiguous regions) to ensure precise localization and successful target convergence, rather than making hasty turns for shorter paths that risk task failure. The official results on the Test-unseen split are publicly available on the R2R-CE leaderboard.

Results on RxR-CE Benchmark (Table \RNum{2}): To verify the generalization capability of DGNav in handling complex, long-horizon instructions, we further conducted evaluations on the RxR-CE dataset. Compared to R2R-CE, RxR-CE features significantly longer trajectories (avg. length increased by $\approx$ 50\%) and more fine-grained instruction descriptions, imposing rigorous demands on the agent's spatiotemporal memory. Table \RNum{2} presents the comparison results against classic methods (Seq2Seq\cite{chang2017matterport3d}, CWP\cite{hong2022bridging}) and recent baselines (ETPNav\cite{an2024etpnav}, StreamVLN\cite{wei2025streamvln}).

Results indicate that DGNav achieves a comprehensive superiority over the ETPNav baseline on the Val-Seen split. Specifically, SR is improved by +1.22\%, and SPL by +0.87\%. More notably, on metrics prioritizing path fidelity—crucial for the RxR task—nDTW and SDTW are boosted by +1.34\% and +1.03\%, respectively. This demonstrates that in familiar environments, DGNav’s dynamic semantic graph can precisely align with the subtle spatial cues described in the instructions.

On the critical Val-Unseen split, DGNav maintains this advantage. Despite the extreme difficulty of this dataset, we achieve consistent improvements in SR and SPL. Particularly on the core metrics measuring “instruction following fidelity”, DGNav again surpasses ETPNav, with nDTW reaching 62.04\% (+0.55\%) and SDTW hitting 44.49\% (+0.57\%). While there is a marginal increase in NE (5.80m to 6.00m), this is not contradictory to the improvement in SR (53.07\% to 53.78\%). This phenomenon typically stems from DGNav performing more local adjustments near the goal area (e.g., fine-tuning based on detailed descriptions), which, while resulting in a slight deviation from the absolute geometric center in Euclidean terms, leads to a more accurate satisfaction of the task completion criteria. Overall, the RxR-CE results compellingly prove that DGNav excels not only at “reaching the destination” but also at “navigating in the prescribed manner”, a direct manifestation of the deep semantic understanding enabled by our dynamic graph approach.

\subsection{Ablation Study}
1) \emph{Analysis of Adaptive Mapping Strategies}

To investigate how to optimally map the predicted scene uncertainty ($\sigma_t$) to the graph construction threshold $\gamma_t$, we compare two distinct linear mapping strategies:
\begin{itemize}
\item Global Linear Mapping: Directly maps the full range of $\sigma_t$ to the $\gamma$ interval.
\item 	Conditional Linear Mapping: Introduces a median value $\sigma_{med}$ as a trigger gate. When $\sigma_t<\sigma_{med}$ (simple scenes), $\gamma_t$ is held at the baseline of 0.5 to maintain stability; the linear mapping to reduce the threshold is triggered only when $ \sigma_t>\sigma_{med}$ (complex scenes).
\end{itemize}

Impact of Triggering Strategy (Global vs. Conditional): Fig. 5 visualizes the histogram of $\gamma_t$ values on the Val-Unseen split for both strategies (excluding the boundary values of 0.25 and 0.5 for visualization clarity). It can be observed that the Global Strategy yields a dispersed distribution of $\gamma$, implying that the model frequently micro-adjusts the threshold even in relatively simple environments, which may introduce unnecessary topological noise. In contrast, the Conditional Strategy exhibits a distinct long-tailed distribution. This pattern indicates that the strategy successfully achieves “Selective Intervention”: the vast majority of simple steps are maintained at the optimal baseline $\gamma_{base}=0.5$ (the dominant peak not shown in the plot), while dynamic adjustments are precisely confined to the tail-end complex scenarios where uncertainty is high.

The quantitative results in Table \RNum{3} further validate the superiority of the Conditional Strategy. Compared to the Global Strategy, the Conditional Strategy achieves a higher SR (58.56\%) and Path SPL (50.08\%). This suggests that establishing a “stability gate” to prevent over-sensitivity in simple scenes, while concentrating resources on addressing sparsity in complex regions, is key to achieving robust navigation. Thus, we adopt the Conditional strategy as our baseline.

\begin{figure}[!t]
\centering
\includegraphics[width=3.5in]{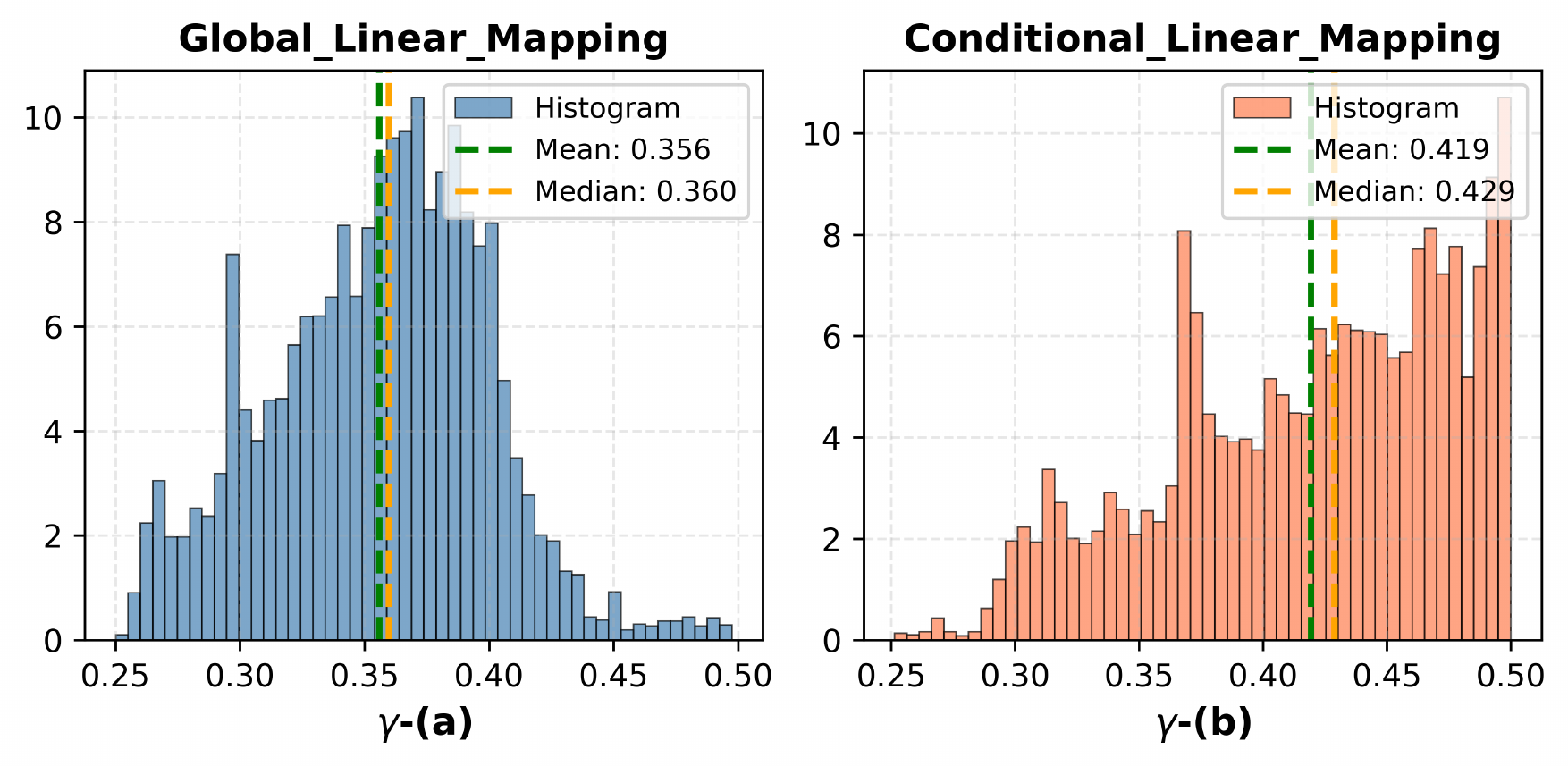}
\caption{Distribution of Dynamic Thresholds ($\gamma_t$) under Different Mapping Strategies on R2R-CE Val-Unseen. (a) Global Linear Mapping exhibits a dispersed distribution, indicating frequent but potentially noisy adjustments. (b) Conditional Linear Mapping shows a long-tailed distribution, reflecting a selective intervention strategy. Note: To clearly visualize the distribution of dynamic adjustments, the dominant peak at the baseline ($\gamma=0.5$) is omitted; data at the lower bound ($\gamma=0.25$) is also excluded to maintain visual symmetry.}
\label{fig_5}
\end{figure}

\emph{\bfseries Choice of Mapping Function}: We validated the superiority of linear mapping against non-linear variants (Sigmoid and Exponential) within the Conditional framework. As illustrated in Fig. 6 and Fig. 7, the Sigmoid function performs the worst; its S-shaped curve suffers from gradient saturation, leading to an overly aggressive distribution ($\gamma\approx0.25$) that generates excessive topological noise. Conversely, the Exponential function exhibits a “Conservative Bias” due to its convex nature, causing “Late Activation” that fails to provide sufficient granularity for moderately complex scenes. In contrast, Linear Mapping achieves the best performance (Table \RNum{3}) by maintaining Constant Sensitivity across the dynamic range. It effectively preserves the entropy of the uncertainty signal, striking an optimal equilibrium that avoids both the “over-reaction” of Sigmoid and the “under-reaction” of Exponential, ensuring precise alignment between topological granularity and environmental complexity.

2) \emph{Effectiveness of Dynamic Graph Construction}

To verify the necessity of dynamically adjusting the threshold $\gamma$, we compared our proposed dynamic scheme against three static threshold settings ($\gamma\in\left\{0.25,\ 0.40,\ 0.50\right\}$) and a random threshold strategy. Table \RNum{4} presents the detailed comparisons on the R2R-CE and RxR-CE validation splits, where $N_{node}$ denotes the average number of generated nodes, reflecting graph density and computational overhead.

Blind Density vs. Intelligent Adaptation: Results indicate that simply lowering the threshold significantly increases graph density but leads to a degradation in navigation performance. This suggests that blindly adding redundant nodes in simple areas not only wastes computational resources but also introduces topological noise that distracts the agent, thereby increasing the difficulty of model learning. Similarly, the Random strategy fails to outperform the baseline, further confirming that the performance gain stems not from random perturbations, but from the precise capture of scene complexity by $\sigma_t$.

\begin{table}[t]
\centering
\caption{ABLATION STUDY ON ADAPTIVE MAPPING STRATEGIES}
\label{tab:ablation}
\renewcommand{\arraystretch}{1.2} 

\resizebox{\columnwidth}{!}{%
\begin{tabular}{c|c|c|ccccc}
\noalign{\hrule height 0.8pt}


\multicolumn{1}{c}{\multirow{2}{*}{Model}} & 
\multicolumn{1}{c}{\multirow{2}{*}{\shortstack{Mapping\\Strategy}}} & 
\multicolumn{1}{c}{\multirow{2}{*}{\shortstack{Mapping\\Function}}} & 
\multicolumn{5}{c}{Val-unseen} \\ \cline{4-8} 

\multicolumn{1}{c}{} & \multicolumn{1}{c}{} & \multicolumn{1}{c}{} & $N_{node}$ & NE$\downarrow$ & OSR$\uparrow$ & SR$\uparrow$ & SPL$\uparrow$ \\ \hline

\multirow{4}{*}{DGNav} 
 & Global & Linear & 27.50 & 4.72 & 63.30 & 57.10 & 49.43 \\
 & \textbf{Condi.} & \textbf{Linear} & \textbf{23.88} & \textbf{4.66} & \textbf{64.82} & \textbf{58.56} & \textbf{50.08} \\
 & Condi. & Sig. & 28.29 & 4.75 & 63.24 & 57.10 & 49.19 \\
 & Condi. & Exp. & 24.73 & 4.73 & 64.55 & 58.02 & 50.03 \\

\noalign{\hrule height 0.8pt}
\end{tabular}%
}
\end{table}

\begin{figure}[!t]
\centering
\includegraphics[width=2.0in]{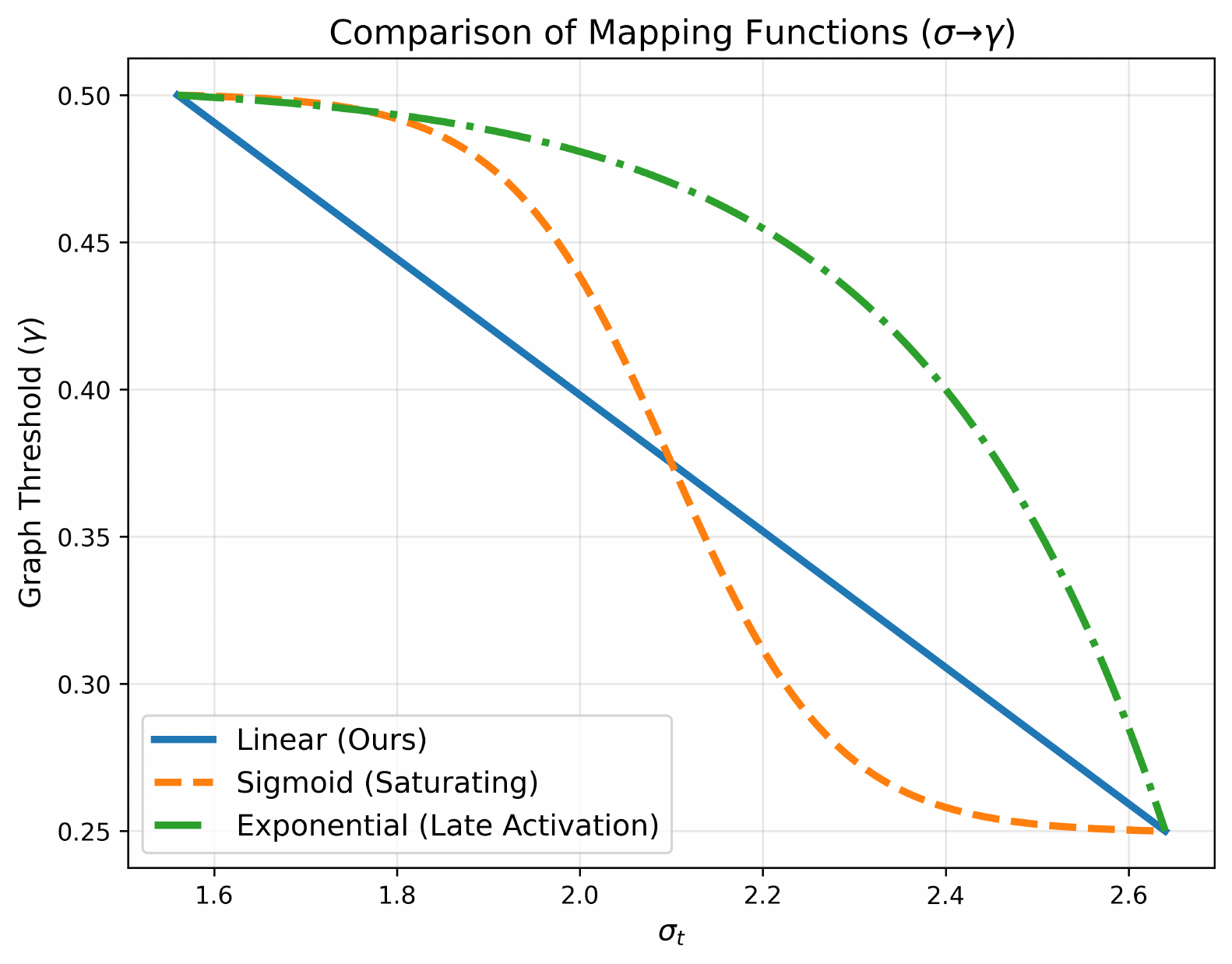}
\caption{Statistical distribution of dynamic thresholds ($\gamma_t$) under non-linear mapping functions on the Val-Unseen split. (a) The Sigmoid mapping exhibits a polarized distribution heavily concentrated near the lower bound ($\gamma\approx0.25$), leading to excessive node generation. (b) The Exponential mapping shows a skewed distribution towards the conservative baseline ($\gamma\rightarrow0.5$), indicating a "late activation" response to scene complexity.}
\label{fig_6}
\end{figure}

\begin{figure}[!t]
\centering
\includegraphics[width=3.5in]{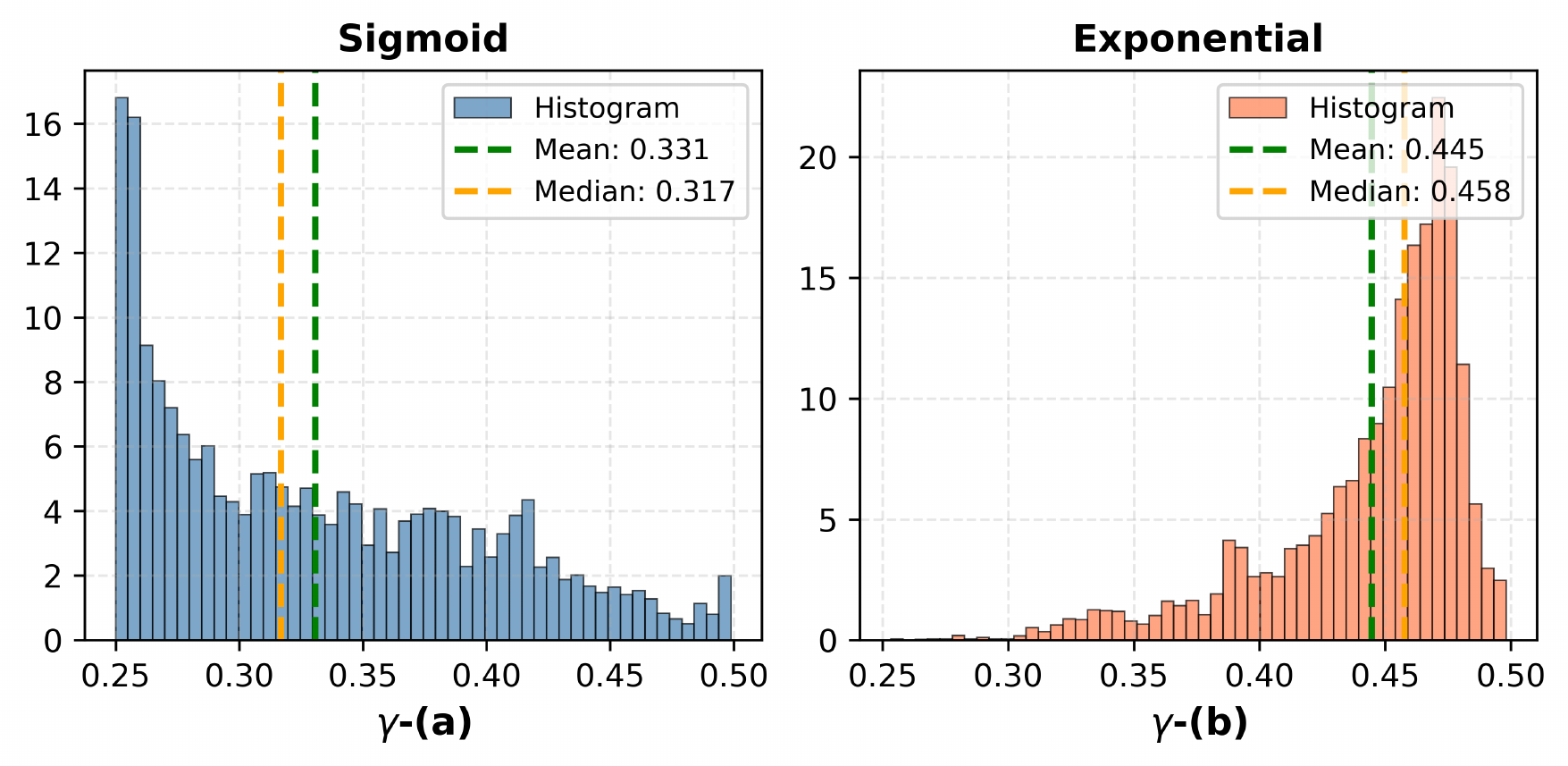}
\caption{Comparison of mapping function shapes ($\sigma_t\rightarrow \gamma_t$). The Linear function provides constant sensitivity, whereas Sigmoid suffers from saturation and Exponential exhibits late activation.}
\label{fig_7}
\end{figure}

\begin{table*}[t]
\centering
\caption{EFFECTIVENESS OF DYNAMIC $\gamma$ STRATEGY}
\label{tab:table4}
\renewcommand{\arraystretch}{1.2} 
\resizebox{\textwidth}{!}{%
\begin{tabular}{c|c|cccccc|cccccc}
\noalign{\hrule height 0.8pt}

\multicolumn{1}{c}{\multirow{2}{*}{Model}} & 
\multicolumn{1}{c}{\multirow{2}{*}{$\gamma$}} & 
\multicolumn{6}{c|}{R2R-CE(Val-unseen)} & 
\multicolumn{6}{c}{RxR-CE(Val-unseen)} \\ \cline{3-14} 

\multicolumn{1}{c}{} & 
\multicolumn{1}{c}{} & 
$N_{node}$ & TL$\downarrow$ & NE$\downarrow$ & OSR$\uparrow$ & SR$\uparrow$ & \multicolumn{1}{c|}{SPL$\uparrow$} & 
$N_{node}$ & NE$\downarrow$ & SR$\uparrow$ & SPL$\uparrow$ & nDTW$\uparrow$ & SDTW$\uparrow$ \\ \hline

\multirow{5}{*}{DGNav} 
 & 0.25 & 31.61 & 11.56 & 4.76 & 62.43 & 56.39 & 49.23 & 46.67 & 6.07 & 53.07 & 44.08 & 61.55 & 43.98 \\
 & 0.40 & 25.90 & 11.46 & 4.75 & 63.19 & 56.66 & 49.03 & 37.40 & 6.02 & 53.35 & 44.07 & 61.53 & 44.21 \\
 & 0.50 & 23.14 & 11.58 & 4.69 & 64.82 & 58.02 & 49.71 & 32.62 & 6.01 & 53.43 & 44.11 & 61.87 & 44.25 \\
 & Random & 26.97 & 11.42 & 4.80 & 62.43 & 56.23 & 48.88 & 39.16 & 6.04 & 53.04 & 43.84 & 60.83 & 44.00 \\
 & Dynamic & 23.88 & 11.64 & \textbf{4.66} & \textbf{64.82} & \textbf{58.56} & \textbf{50.08} & 33.98 & \textbf{6.00} & \textbf{53.78} & \textbf{44.37} & \textbf{62.04} & \textbf{44.49} \\

\noalign{\hrule height 0.8pt}
\end{tabular}%
}
\end{table*}

In contrast, our Dynamic strategy achieves the best performance across both datasets. Specifically, on R2R-CE, the Dynamic strategy leads in all metrics (NE, OSR, SR, SPL). Notably, this improvement is achieved with minimal additional structural complexity. Since the computational load of the Graph Transformer scales with the number of nodes, the marginal increase in average node count ($N_{node}$ increases by only $\approx$ 0.4) implies that DGNav incurs negligible runtime overhead compared to the baseline, while yielding significant performance gains. On RxR-CE, this advantage is even more pronounced. The Dynamic strategy not only tops the SR and SPL metrics but also significantly outperforms static and random schemes in nDTW and SDTW, which measure path fidelity. This confirms that the dynamic topology better accommodates the reliance on fine-grained spatial landmarks in long-horizon navigation.

To provide a more intuitive validation of this mechanism, we visualize the graph generation process across the first three navigation steps of the same episode in Fig. 8. From the top-down view, it is clearly observable that at the initial pose (Step 0), facing multiple navigable directions, the baseline model is constrained by a fixed threshold and generates only 3 ghost nodes. In contrast, our dynamic strategy perceives the environmental complexity and actively lowers the $\gamma$ value, generating 4 ghost nodes to cover more potential paths. This advantage persists through the second and third steps; notably, in the complex multi-junction scenario at Step 2, the dynamic strategy generates 4 additional ghost nodes compared to the fixed baseline. This significant node increment strongly confirms that DGNav effectively aligns topological granularity with the immediate environmental complexity through “densification on demand”.

3) \emph{Impact of Multi-modal Dynamic Edge Fusion}

Next, we investigate the contribution of different modal edge weights in the dynamic graph. To evaluate the specific impacts of $\mathbf{E}_{\mathbit{sem}}$ and $\mathbf{E}_{\mathbit{inst}}$ on graph connectivity, we conducted component ablation studies on the R2R-CE Val-Unseen dataset. Considering the computational cost of re-training, this part of the ablation is primarily performed on R2R-CE. All experiments were conducted with the Adaptive $\gamma$ strategy enabled.

\begin{figure}[!t]
\centering
\includegraphics[width=3.5in]{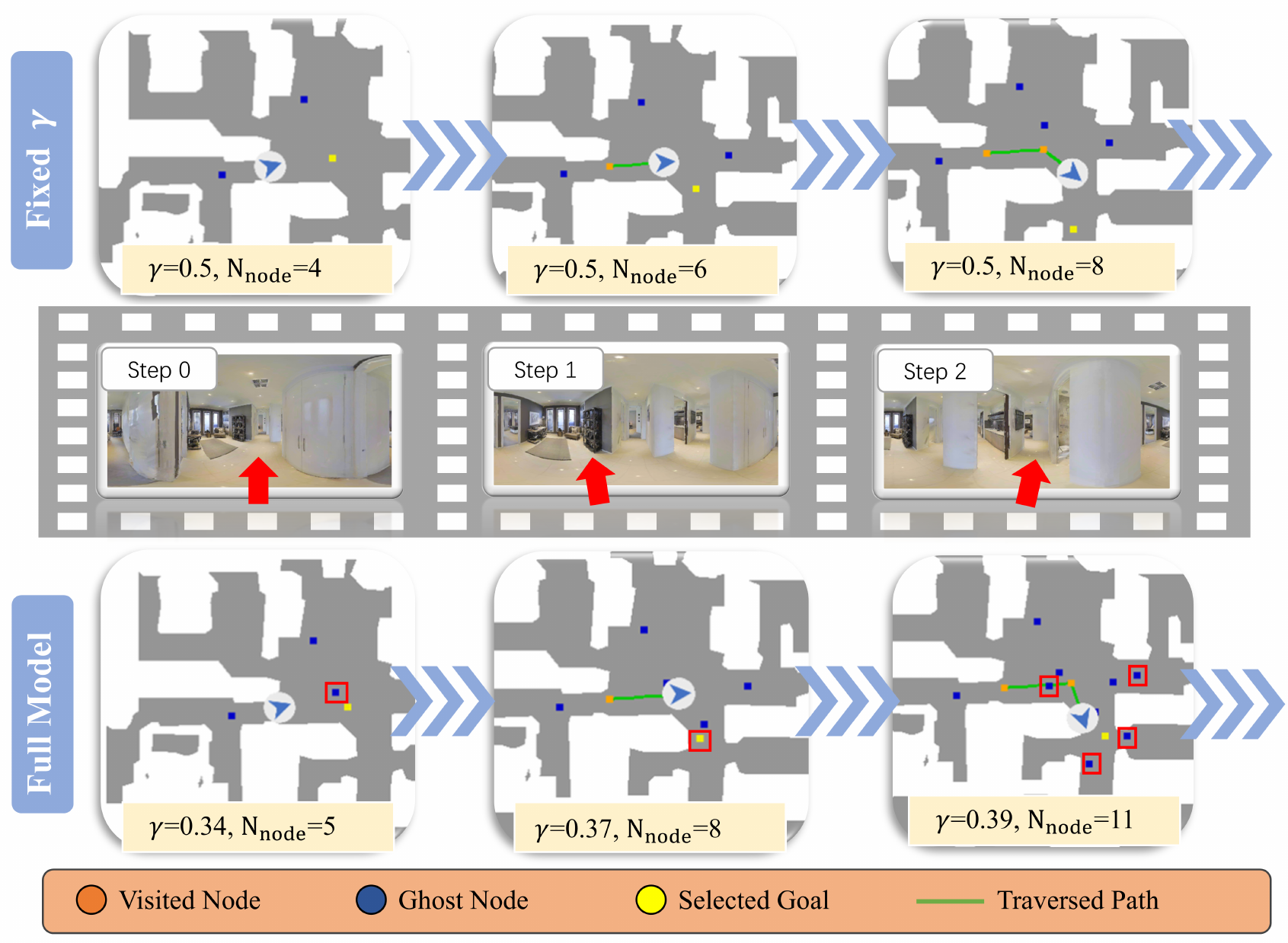}
\caption{Visualization of Scene-Aware Adaptive Graph Construction. Comparison of graph generation in a complex environment within the same episode. (Top) Fixed $\gamma$: Constrained by a static threshold, the baseline generates a sparse set of ghost nodes. (Bottom) DGNav: The Scene-Aware Adaptive Strategy detects high local uncertainty and dynamically lowers the threshold, triggering a dense generation of Ghost Nodes ($N_{node}\uparrow$).}
\label{fig_8}
\end{figure}

\begin{table}[t]
\centering
\caption{ABLATION STUDY ON DYNAMIC EDGE FUSION MODULES}
\label{tab:table5}
\renewcommand{\arraystretch}{1.2} 
\resizebox{\columnwidth}{!}{%
\begin{tabular}{c|cc|cccccc}
\noalign{\hrule height 0.8pt}

\multicolumn{1}{c}{\multirow{2}{*}{Model}} & 
\multicolumn{1}{c}{\multirow{2}{*}{\boldmath$E_{sem}$}} & 
\multicolumn{1}{c}{\multirow{2}{*}{\boldmath$E_{inst}$}} & 
\multicolumn{6}{c}{Val-unseen} \\ \cline{4-9} 

\multicolumn{1}{c}{} & 
\multicolumn{1}{c}{} & 
\multicolumn{1}{c}{} & 
Steps & TL$\downarrow$ & NE$\downarrow$ & OSR$\uparrow$ & SR$\uparrow$ & SPL$\uparrow$ \\ \hline

\multirow{4}{*}{DGNav} 
 & - & - & 80.04 & 11.44 & 4.76 & 63.62 & 56.82 & 48.74 \\
 & $\surd$ & - & 71.00 & 10.38 & 4.80 & 61.94 & 55.85 & 49.52 \\
 & - & $\surd$ & 77.38 & 11.05 & 4.88 & 61.77 & 56.01 & 48.71 \\
 & $\surd$ & $\surd$ & 80.36 & 11.64 & \textbf{4.66} & \textbf{64.82} & \textbf{58.56} & \textbf{50.08} \\

\noalign{\hrule height 0.8pt}
\end{tabular}%
}
\end{table}

As shown in Table \RNum{5}, relying solely on geometric distance as a baseline limits the model’s semantic understanding of the environment. Interestingly, when introducing $\mathbf{E}_{\mathbit{sem}}$ alone, we observe a significant drop in average Steps and TL and an increase in SPL, but a slight decline in SR. This suggests that pure visual similarity connections make the agent “aggressive and efficient”, prone to taking visual shortcuts; however, without instruction constraints, these shortcuts sometimes lead to erroneous navigation decisions. In contrast, the Full Model (DGNav), which fuses both $\mathbf{E}_{\mathbit{sem}}$ and $\mathbf{E}_{\mathbit{inst}}$, achieves the best performance, with SR jumping to 58.56\% and SPL reaching 50.08\%. This demonstrates the synergy between the two modalities: $\mathbf{E}_{\mathbit{sem}}$ identifies potential semantic shortcuts to enhance efficiency, while $\mathbf{E}_{\mathbit{inst}}$ acts as a “task-alignment filter” to verify that these shortcuts comply with the instruction descriptions. The combination allows the agent to optimize path planning while maintaining a high success rate.

\begin{figure*}
    \centering
    \includegraphics[width=0.7\linewidth]{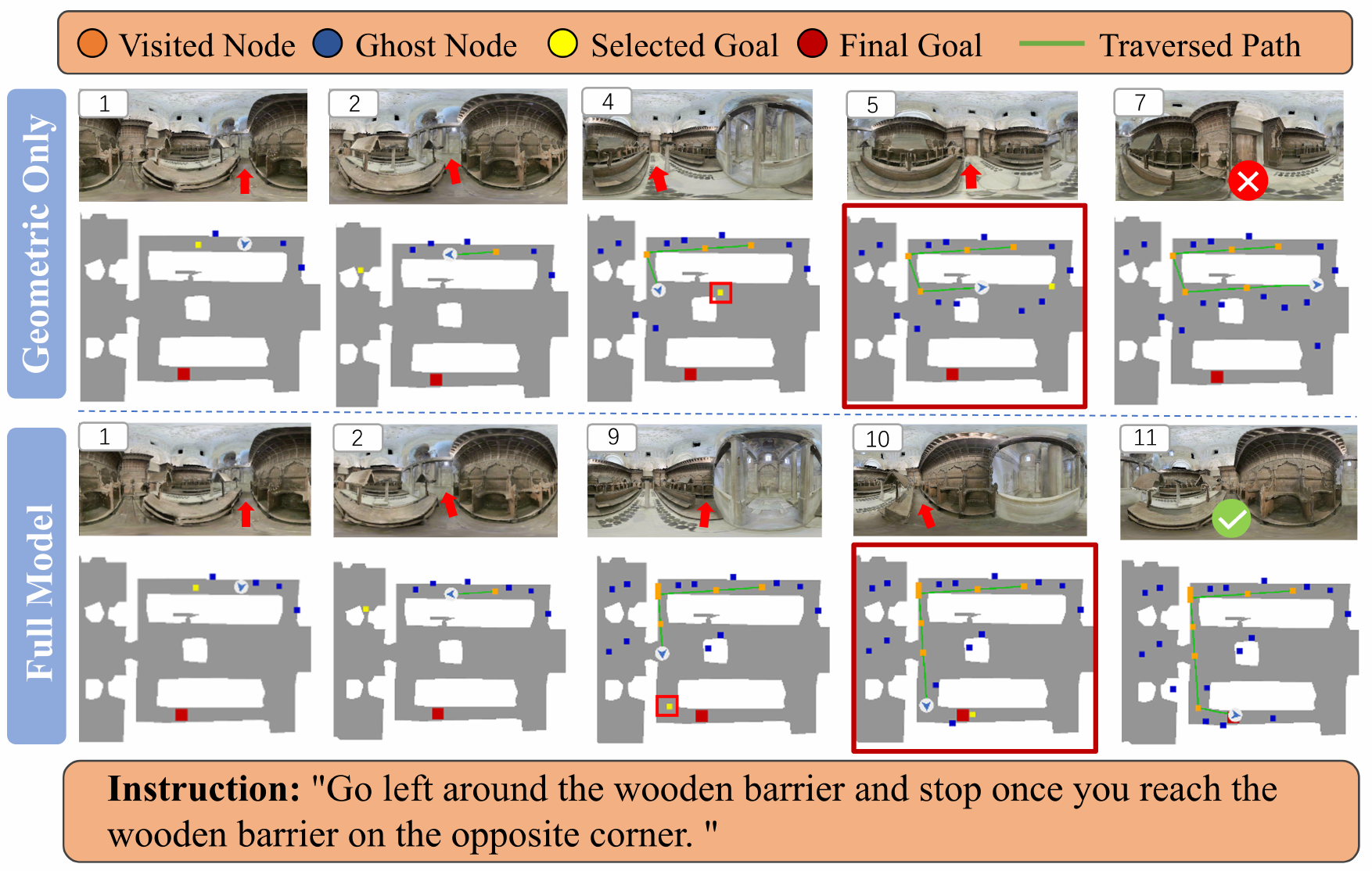}
    \caption{Qualitative Analysis: Static Geometric Priors vs. Dynamic Semantic Weights. (Top) Geometric Only: Lacking semantic filtering, the agent is misled by static geometric priors and selects the incorrect turn solely due to its proximity. (Bottom) DGNav: Empowered by Multi-modal Dynamic Edge Fusion, the agent effectively integrates visual and linguistic cues to suppress the weight of the incorrect path. It correctly identifies the instructional intent, ignores the geometric distractor, and boldly explores the open space to successfully reach the target.}
\end{figure*}

To further qualitatively validate the effectiveness of the Multi-modal Dynamic Edge Fusion module, we present a complete navigation episode in Fig. 9. In this scenario, the agent faces two consecutive left-turn intersections, while the instruction explicitly requires it to “reach the wooden barrier” before turning. Lacking semantic perception, the Geometric-Only baseline is misled by static geometric priors: it executes the turn prematurely upon merely “seeing” the barrier rather than truly “reaching” it, causing a complete deviation from the target path. This reveals a fundamental limitation of geometry-dominated planners. In contrast, DGNav, leveraging dynamic edge weights, effectively fuses visual and linguistic cues to suppress the connectivity of the incorrect turn. In this large-scale environment, it overcomes the geometric bias and makes a decision that aligns better with the instructional semantics, boldly choosing a farther but correct exploration path to successfully reach the destination. This case study intuitively demonstrates the critical role of dynamic edge fusion in correcting geometric biases and enhancing instruction adherence.

4) \emph{Exploration of Advanced Training Strategies}

To further enhance semantic dependency, we investigated three training variants: (1) \emph{Node Gating}: A learnable MLP head to filter nodes based on predicted topological importance; (2) \emph{Geometric Dropout}: Randomly masking the geometric stream to prevent overfitting to explicit distances; (3) \emph{Geometric Annealing}: A curriculum strategy where dropout linearly increases to stabilize the transition from geometric to semantic reliance.

Analysis of Results (as shown in Table \RNum{6}):
\begin{itemize}
\item \emph{Redundancy of Node Gating}: Introducing gating causes a slight drop in SPL. This suggests that our core Adaptive Threshold ($\gamma$) is already sufficient for density control, and additional filtering likely removes necessary stepping-stone nodes, causing detours.
\item \emph{Enhanced Exploration via Geometric Dropout}: Enabling dropout significantly increases TL and OSR. This indicates that reducing geometric priors “liberates” exploration capabilities, encouraging the agent to venture into unknown areas, albeit at the cost of path efficiency.
\item \emph{Performance Upper Bound via Geometric Annealing}: The annealing strategy achieves the highest SR and lowest NE. This confirms that balancing “geometry-guided convergence” and “semantics-driven localization” is crucial. Although its SPL is lower than standard DGNav, this strategy effectively curbs risky geometric shortcuts, steering the agent towards conservative but accurate paths—a promising direction for safety-critical scenarios.
\end{itemize}

Considering the balance between computational efficiency and navigation performance, we retained the standard DGNav for our main experiments. However, these explorations compellingly demonstrate that further decoupling geometric dependency through implicit training strategies is an effective pathway to enhancing the robustness of VLN agents.

\begin{table}[t]
\centering
\caption{Analysis of Advanced Training Strategies}
\label{tab:table6}
\renewcommand{\arraystretch}{1.2}
\resizebox{\columnwidth}{!}{%
\begin{tabular}{c|ccc|ccccc}
\noalign{\hrule height 0.8pt}

\multicolumn{1}{c}{\multirow{2}{*}{Method}} & 
\multicolumn{1}{c}{\multirow{2}{*}{Gate}} & 
\multicolumn{1}{c}{\multirow{2}{*}{Drop}} & 
\multicolumn{1}{c}{\multirow{2}{*}{Ann}} & 
\multicolumn{5}{c}{Val-unseen} \\ \cline{5-9}

\multicolumn{1}{c}{} & 
\multicolumn{1}{c}{} & 
\multicolumn{1}{c}{} & 
\multicolumn{1}{c}{} & 
TL & NE$\downarrow$ & OSR$\uparrow$ & SR$\uparrow$ & SPL$\uparrow$ \\ \hline

\multirow{4}{*}{DGNav} 
 & - & - & - & 11.64 & 4.70 & 64.82 & 58.56 & \textbf{50.08} \\
 & $\surd$ & - & - & 12.35 & 4.77 & 64.82 & 58.40 & 49.28 \\
 & $\surd$ & $\surd$ & - & 14.08 & 4.71 & 66.83 & 58.56 & 47.21 \\
 & $\surd$ & $\surd$ & $\surd$ & 13.55 & \textbf{4.60} & \textbf{67.43} & \textbf{59.11} & 48.16 \\

\noalign{\hrule height 0.8pt}
\end{tabular}%
}
\end{table}

\section{CONCLUSION}
In this paper, we address the “Granularity Rigidity” dilemma inherent in existing topological planning methods for Vision-Language Navigation in Continuous Environments (VLN-CE). We propose DGNav, a framework that breaks the constraints of fixed construction thresholds and sole geometric metrics by introducing a dynamic topological perception mechanism. Furthermore, the Dynamic Graph Transformer reconstructs graph connectivity by fusing visual semantics, instruction relevance, and geometric constraints, effectively suppressing topological noise and enhancing the fidelity of path planning. Extensive experiments on R2R-CE and RxR-CE benchmarks demonstrate that DGNav achieves strong performance against strong baselines, exhibiting particular strengths in path efficiency and robust spatiotemporal alignment.

\bibliographystyle{ieeetr}
\bibliography{reference}

\end{document}